\documentclass[10pt,twocolumn,letterpaper]{article}

\usepackage[pagenumbers]{cvpr} 

\usepackage{xcolor}
\usepackage{soul}

\usepackage{booktabs}
\usepackage{siunitx}
\usepackage{adjustbox}
\usepackage{booktabs,multirow,siunitx,adjustbox,threeparttable,colortbl}
\usepackage{cuted}  
\usepackage{float}  

\usepackage{url}
\usepackage{algorithm, algpseudocode}
\usepackage{amssymb}  
\usepackage{floatflt}

\usepackage{kotex}
\usepackage{amsmath}
\usepackage{amsfonts}
\usepackage{subcaption}  
\usepackage{multirow}
\usepackage{booktabs}
\usepackage{threeparttable}

\usepackage{afterpage}
\usepackage{adjustbox}  
\usepackage{subcaption}
\usepackage{graphicx,wrapfig,caption}
\usepackage{booktabs,multirow,makecell}
\usepackage{tabularx}
\newcolumntype{C}{>{\centering\arraybackslash}X}
\usepackage{siunitx}
\usepackage{bigstrut}

\newcommand{\lift}{\textbf{LIFT}}
\newcommand{\flift}{\textbf{LI}near \textbf{F}i\textbf{T}ting-based distillation}
\newcommand{\blift}{LInear FiTting-based distillation}
\newcommand{\place}{\textbf{PLACE}}
\newcommand{\fplace}{\textbf{P}iecewise \textbf{L}ocal \textbf{A}daptive \textbf{C}oefficient \textbf{E}stimation}
\newcommand{\bplace}{Piecewise Local Adaptive Coefficient Estimation}

\definecolor{teacherbg}{gray}{0.92}  

\renewcommand{\paragraph}[1]{\vspace{.5em}\noindent\textbf{#1}}

\DeclareMathOperator*{\argmin}{arg\,min}

\definecolor{cvprblue}{rgb}{0.21,0.49,0.74}
\usepackage[pagebackref,breaklinks,colorlinks,allcolors=cvprblue]{hyperref}

\def\paperID{} 
\def\confName{CVPR}
\def\confYear{2026}

\title{LIFT and PLACE: A Simple, Stable, and Effective Knowledge Distillation Framework for Lightweight Diffusion Models}

\author{
Hyunsoo Han, Sangyeop Yeo, Jaejun Yoo\\
Ulsan National Institute of Science and Technology (UNIST) \\
{\tt\small \{hyuns, sosick377, jaejun.yoo\}@unist.ac.kr}}

\begin{document}
\maketitle


\begin{abstract}
We demonstrate that in knowledge distillation for diffusion models, the teacher network’s highly complex denoising process—stemming from its substantially larger capacity—poses a significant challenge for the student model to faithfully mimic. To address this problem, we propose a coarse-to-fine distillation framework with \flift~(\lift) and \fplace~(\place). First, \lift~decomposes the objective into a ``coarse'' alignment and a ``fine'' refinement. The student is then trained on coarse alignment before proceeding to hard refinement. Second, \place~extends LIFT to address spatially non-uniform errors by partitioning outputs into error-based groups, providing locally adaptive guidance.
Our experiments show that \lift~and \place~is effective across diffusion spaces (image/latent), backbones (U-Net/DiT), tasks (unconditional/conditional), datasets, and even extends to flow-based models such as MMDiT (SD3).
Furthermore, under extreme compression with a 1.3M-parameter student (only 1.6\% of the teacher),  conventional KD fails to provide sufficient guidance for stable training, with FID scores often degrading to 50--200+, but our method remains stably convergent and achieves an FID of 15.73. Our project page is available at \href{https://hyun-s.github.io/LIFT_PLACE_site/}{here}.
\end{abstract}    
\section{Introduction}
\label{sec:intro}


\begin{figure}[t]
  \centering
   \includegraphics[width=1.0\linewidth]{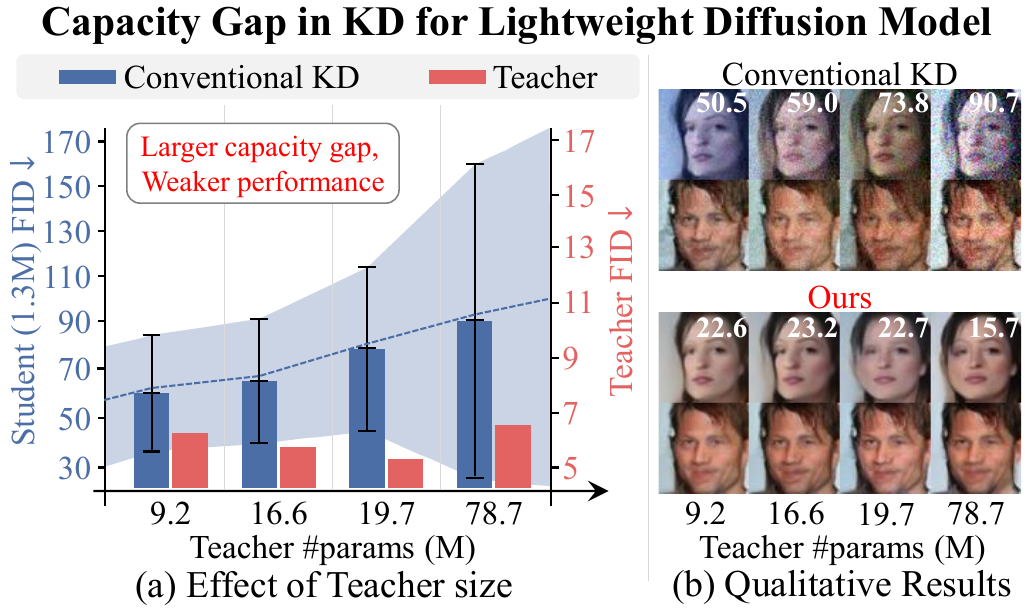}
  \caption{
  Impact of teacher network scale on distilling diffusion models into students. (a) illustrates the average performance over five runs of knowledge distillation. The blue bar indicates the average performance of the student network, and the shaded blue region indicates its standard deviation. As the capacity (model size) gap increases, the student's performance not only degrades but also becomes increasingly unstable, as reflected by the larger standard deviation. (b) shows that image quality deteriorates with larger teachers, whereas our method preserves fidelity.
  }
  \label{fig:teaser}
\end{figure}
As the saying goes, ``\emph{A great scholar is not necessarily a great teacher}" indicating that being highly knowledgeable does not always translate into effective teaching. This notion carries over to knowledge distillation (KD): as the capacity of a teacher network increases, a smaller student network---with limited capacity and representational power---often struggles to effectively mimic the teacher's behavior. 
A growing body of work~\citep{cho2019efficacy,mirzadeh2020improved, huang2022knowledge,qian2025good,li2025dual} has shown that KD becomes less effective when the capacity gap widens, and that moderate teachers often yield better transfer than overly large ones. 


This challenge is magnified in diffusion models~\citep{fang2025tinyfusion,kim2024bk,kim2025random,xiang2025dkdm,zhang2024laptop,lee2024koala}. Their iterative denoising nature forces the student to learn a deep hierarchy of noise prediction behaviors, making the distillation process particularly demanding. Empirically, as shown in \cref{fig:teaser}, when the teacher-student capacity gap increases, the distilled student not only underperforms but also displays higher performance variance across runs, suggesting that the training process becomes unstable. Yet, despite the rapid progress in lightweight diffusion models, to the best of our knowledge, the role of capacity gap remains largely underexplored in this setting.

\begin{figure}[t]
  \centering
  \vspace{-8pt}
  \includegraphics[width=1\linewidth]{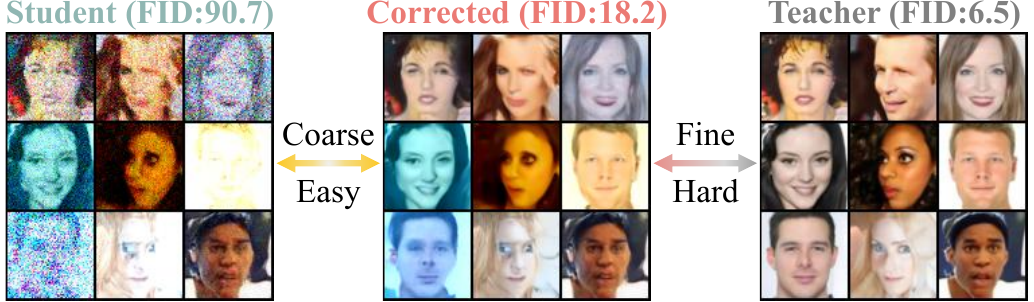}
  \captionsetup{width=1\linewidth, 
  }
  \caption{ 
  Regression-based correction analysis. At each time step $t$, we estimate regression coefficients from teacher-student pairs and affine-correct the student output, following \cref{alg:linear_correction}.
  Qualitative samples are shown below. Left: student; middle: corrected as in \cref{alg:linear_correction}; right: teacher. The corrected sampler yields more faithful images. However, a performance gap between the corrected output and the teacher remains, especially for fine and hard details.}
  \label{fig:obs}
\end{figure}

A common workaround is to deliberately  weaken the teacher---using smaller capacities~\citep{mirzadeh2020improved,son2021densely}~or intentionally suboptimal teachers~\citep{cho2019efficacy,wang2022efficient}. While this can simplify optimization, it inevitably restricts the richness of the knowledge available to the student. What we ultimately desire is the opposite: reliably transmitting the full expressiveness of a strong teacher into a student without sacrificing information. Then, ``\textit{how?}"
Our investigation begins with a simple yet revealing experiment. For each sample, we independently apply linear regression at each denoising time step to match the teacher’s and student’s predicted noise (\cref{alg:linear_correction}). This operation aligns only the low-order statistical moments of their output distributions---essentially a coarse matching of mean and variance---while ignoring any higher-order structure. Surprisingly, we find that even this crude operation drastically stabilizes and improves student training compared to conventional KD (\cref{fig:obs}). 


However, this analysis also exposed an inherent limitation. While linear regression resolves the coarse, easy-to-learn discrepancies, it \textit{leaves untouched the fine-grained, non-linear structure} that characterizes a strong diffusion teacher, which is difficult for lightweight students.
More importantly, performing a regression \textit{for each sample at every denoising step} would require access to the teacher network and the target sample even at inference time, making it fundamentally unsuitable as a real distillation strategy. 

What this experiment provides, therefore, is not a deployable method but a diagnostic lens. It suggests that the distillation error naturally decomposes into two components:
(1) ``Coarse-Easy" errors—low-order statistical mismatches that students can readily learn, and
(2) ``Fine-Hard" errors—non-linear residuals encoding the richer structure of the teacher.
Crucially, forcing the student to optimize both simultaneously destabilizes KD, especially under large capacity gaps. This motivates the need to amortize the beneficial effect of coarse alignment into a practical training objective that operates globally across the dataset---without requiring per-sample regression during inference.


Our method, \blift~(\lift), formalizes this \textit{Coarse-to-Fine} training strategy by turning the diagnostic insights above into a practical and amortized training objective. Instead of per-sample regressions, LIFT parameterizes the KD objective with a single global linear regression and explicitly decomposes the teacher–student discrepancy into two parts: (1) a coarse alignment term that regularizes the regression coefficients to capture the easy, low-order statistical structure, and (2) a fine-detail refinement that minimizes the remaining non-linear residual. As training progresses, an adaptive weighting mechanism gradually shifts the emphasis from coarse alignment to residual refinement, enabling the student to first stabilize around the global statistics before learning the teacher’s fine-grained behavior. This amortization allows the benefit of the probe experiment---stability from coarse alignment---to be realized consistently across all data and all denoising steps.

Although LIFT captures global alignment, the residual error between teacher and student is not spatially uniform. Certain regions of the output exhibit larger discrepancies and require stronger corrective refinement. To address this spatial heterogeneity, we introduce PLACE. PLACE extends LIFT by providing locally adaptive guidance: it partitions the output into groups according to distillation difficulty, measured by residual magnitude, and applies LIFT independently within each group. This piecewise refinement preserves LIFT's simplicity while allowing the model to focus additional capacity precisely where the student struggles most without adding parameters or inference overhead.
When the capacity gap is large, where conventional KD methods often fail to converge, we show that our approach consistently achieves stable training and strong performance across diffusion spaces (image/latent), backbones (U-Net/DiT), tasks (unconditional/conditional), datasets, and even generative paradigms beyond diffusion (i.e., flow-matching). Prior distillation studies in generative models have typically been limited to a single backbone, task, diffusion space, or generative formulation. To the best of our knowledge, this is the first work to systematically evaluate distillation across such diverse settings and demonstrate consistent improvements throughout. 
We summarize our contributions as follows:

\begin{itemize} 
    \item {We highlight the performance drop caused by the capacity gap between the teacher and student networks in knowledge distillation for diffusion models.}
    \item We propose LIFT that decomposes the KD objective into ``Coarse-Easy" and ``Fine-Hard" components, and PLACE, which provides locally adaptive refinement by error-based grouping.
    \item 
    Our experiments demonstrate that our framework is architecture-agnostic, task-agnostic, and introduces no additional parameters or inference overhead.
\end{itemize}
%

\section{Background}
Knowledge distillation (KD) for compressed diffusion models \citep{kim2024bk,lee2024koala,fang2025tinyfusion,zhang2024laptop,gupta2024progressive} aims to minimize the distance between the outputs $\epsilon$ or intermediate feature of the teacher $G^{\mathcal{T}}$ and student $G^{\mathcal{S}}$ networks, enabling the student to mimic the teacher. Output-level KD matches the output noise $\epsilon$ of $G^{\mathcal{T}}$ and $G^{\mathcal{S}}$ at each denoising time step:
\begin{align}
\mathcal L_{\text{OutKD}} &= \mathbb E_{x,t}[||\epsilon^{\mathcal T}(x_t,t;\theta^{\mathcal T})-\epsilon^{\mathcal S}(x_t,t;\theta^{\mathcal S})||^2_2]
\label{eq:output}
\end{align}
where $x_t$ is a noisy input of the diffusion process according to sampled $\epsilon\sim\mathcal N(0,I)$ and $t\sim U(1,T)$, and $\theta^{\mathcal T}$ and $\theta^{\mathcal S}$ are the teacher and student parameters, respectively. 

In contrast, feature-level KD matches the intermediate features $f_i$ of the $i$-th layers of U-Net~\citep{ronneberger2015u}~or DiT~\citep{dit}~architectures during each denoising step:
\begin{align}
\mathcal L_{\text{FeatKD}} &= \mathbb E_{x,t}[||f_i^{\mathcal T}(x_t,t;\theta^{\mathcal T}){-}r(f_i^{\mathcal S}(x_t,t;\theta^{\mathcal S}))||^2_2] 
\label{eq:feat}
\end{align}
where $r$ is a regressor (e.g., $1{\times}1$ convolution) employed to align the feature dimensions. Throughout, we abbreviate output-level KD and feature-level KD as \textbf{OutKD} and \textbf{FeatKD}, respectively, and denote their losses by~\cref{eq:output,eq:feat}. Unless otherwise noted, the corresponding KD loss is combined with the diffusion loss. For brevity, we write $\epsilon^{\mathcal T}(\cdot)$, $\epsilon^{\mathcal S}(\cdot)$, $f_i^{\mathcal T}(\cdot)$ and $f_i^{\mathcal S}(\cdot)$  simplify as $\epsilon^{\mathcal{T}}$, $\epsilon^{\mathcal{S}}$, $f_i^\mathcal T$ and $f_i^\mathcal S$, and we omit
the expectation $\mathbb E_{x,t}[\cdot]$.


\begin{algorithm}[t]
\footnotesize
\captionsetup{font=footnotesize}
\caption{\footnotesize (Per Sample) Regression-based Correction Analysis}
\label{alg:linear_correction}
\begin{algorithmic}[1]

    \State \textbf{Input:} Teacher model $\epsilon^T$, Student model $\epsilon^S$
    \State \textbf{Output:} Corrected sample $x_0 \in \mathbb R^{C\times H\times W}$
    \State $x_T \sim \mathcal{N}(0, \mathbf{I})$
    \For{$t = T, T-1, \dots, 1$}
        \State $\epsilon_t^\mathcal T \gets \epsilon^\mathcal T(x_t, t)$,\quad $\epsilon_t^\mathcal S \gets \epsilon^\mathcal S(x_t, t)$
        \State $(\hat{\beta}_0, \hat{\beta}_1) \gets \text{OLS}(\epsilon_t^\mathcal T, \epsilon_t^S)$ \Comment{$\hat \beta_{0}\in \mathbb R^{1},\hat \beta_{1}\in \mathbb R^{1}$}
        \State $\hat{\epsilon}_t^S \gets \hat{\beta}_0 + \hat{\beta}_1 \cdot \epsilon_t^S$ \Comment{Linear correction}
        \State $z \sim \mathcal{N}(0, \mathbf{I})$ \textbf{if} $t > 1$ \textbf{else} $z = 0$
        \State $x_{t-1} \gets \frac{1}{\sqrt{\alpha_t}} \left( x_t - \frac{1-\alpha_t}{\sqrt{1-\bar{\alpha}_t}} \hat{\epsilon}_t^S \right) + \sigma_t z$
    \EndFor
    \State \Return $x_0$
\end{algorithmic}
\end{algorithm}

\section{Distillation Error Analysis}\label{sec:analysis} 
To understand the training instability caused by the large capacity gap, we first provide a comprehensive analysis of the distillation error. Our analysis reveals two fundamental properties of this error, which motivate our method: we first examine its statistical components and investigate where this error appears, revealing its spatial non-uniformity.

\subsection{Decomposing Distillation Error}\label{sec:decompose}
We start by treating the student and teacher model's predictions $\epsilon^{\mathcal{S}}$, $\epsilon^{\mathcal{T}}$ at each timestep as distinct distributions. These distributions are characterized by their statistical moments, such as mean and variance. Low-order (i.e., first and second) moments provide summarized information on the given distribution but capture less detailed information. High-order moments capture fine-grained characteristics of a distribution that low-order moments do not. 

To explore this, as shown in \cref{fig:obs} and \cref{alg:linear_correction}, we approximate low-order moment alignment using linear regression, ${\epsilon}^{\mathcal{T}} \approx \beta_0 + \beta_1 \cdot {\epsilon}^{\mathcal{S}}$. 
This regression can be solved by ordinary least squares (OLS) per sample
:
\begin{gather}
\beta_1=\text{Cov}[\epsilon^\mathcal T,\epsilon^\mathcal S]/\text{Var}[\epsilon^\mathcal S],\quad \beta_0=\mathbb E[\epsilon^\mathcal T]-\beta_1 \mathbb E[\epsilon^\mathcal S]. \label{eq:ols}
\end{gather}
Applying this correction, $\hat{\epsilon}^{\mathcal{S}} = \beta_0 + \beta_1 \cdot {\epsilon}^{\mathcal{S}}$, leads to significant performance gains, even when the student's weights remain frozen. Nevertheless, a performance gap between the student and teacher still remains. This implies that the errors uncorrected by moment alignment are responsible for this remaining gap. Last but not least, applying this correction during inference is practically infeasible; estimating $\beta_0$ and $\beta_1$ per sample requires teacher model's inference at every timestep to obtain $\epsilon^{\mathcal{T}}$, which completely defeats the efficiency purpose of distillation.

This analysis allows us to decompose the distillation error into two types:
(1) ``Coarse-Easy’’ Errors: the statistical discrepancy between the low-order moments of the two distributions. Minimizing these errors makes a stable denoising process in diffusion models.
(2) ``Fine-Hard’’ Errors: the remaining, complex discrepancies that are not captured by the first and second-order moments. These errors manifest as the remaining performance gap and a lack of fine-grained details.
This error decomposition allows us to move beyond a monolithic KD objective and address these errors separately, as detailed in \cref{sec:LIFT}.




\begin{figure}[t]
  \centering
  \begin{subfigure}[t]{0.325\linewidth}
    \centering
    \includegraphics[width=\linewidth,]{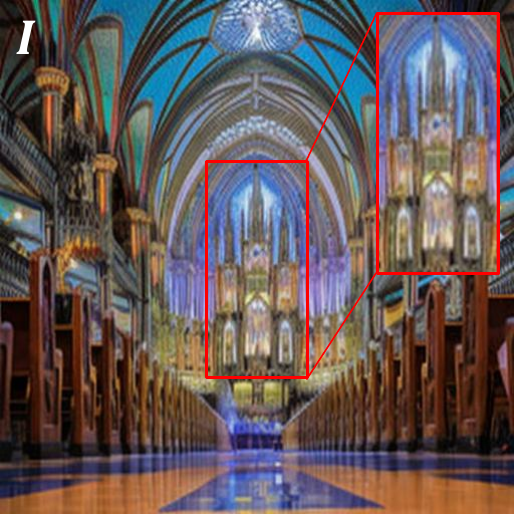}
    \captionsetup{width=1\linewidth}
    \caption{Input image.}
    \label{fig:error_map_c}
  \end{subfigure}
  \begin{subfigure}[t]{0.325\linewidth}
    \centering
    \includegraphics[width=\linewidth]{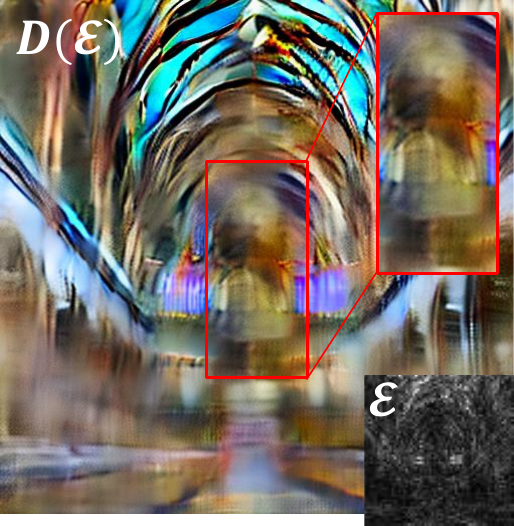}
    \captionsetup{width=1\linewidth}
    \caption{Early iteration, 1k.}
    \label{fig:error_map_a}
  \end{subfigure}
  \hfill
  \begin{subfigure}[t]{0.325\linewidth}
    \centering
    \includegraphics[width=\linewidth]{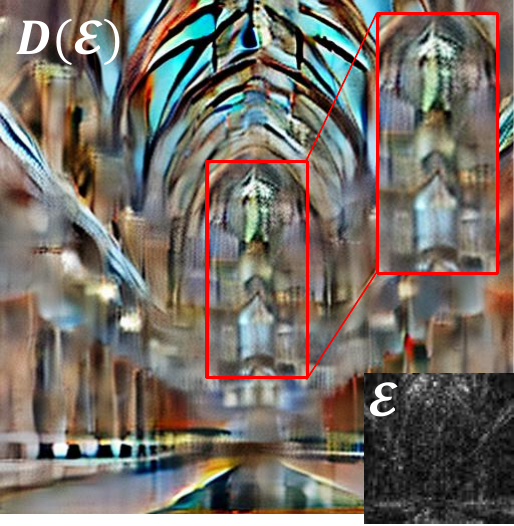}
    \captionsetup{width=1\linewidth}
    \caption{After training, 50k.}
    \label{fig:error_map_b}
  \end{subfigure}
  \captionsetup{width=1\linewidth}
  \caption{
  Visualization of (a) input image, latent error map $\mathcal{E}=|\epsilon^{\mathcal T}-\epsilon^{\mathcal S}|$ and decoded error map ${D}(\mathcal{E})$ of BK-Tiny-v2 at (b) early iteration and (c) after training. 
  The error is spatially non-uniform and evolves over the training process.
  }

  \label{fig:error_map}
\end{figure}
\subsection{Spatially Non-uniform Distillation Error}\label{sec:err}
\begin{figure*}[t]
  \centering
    \includegraphics[width=1\textwidth]{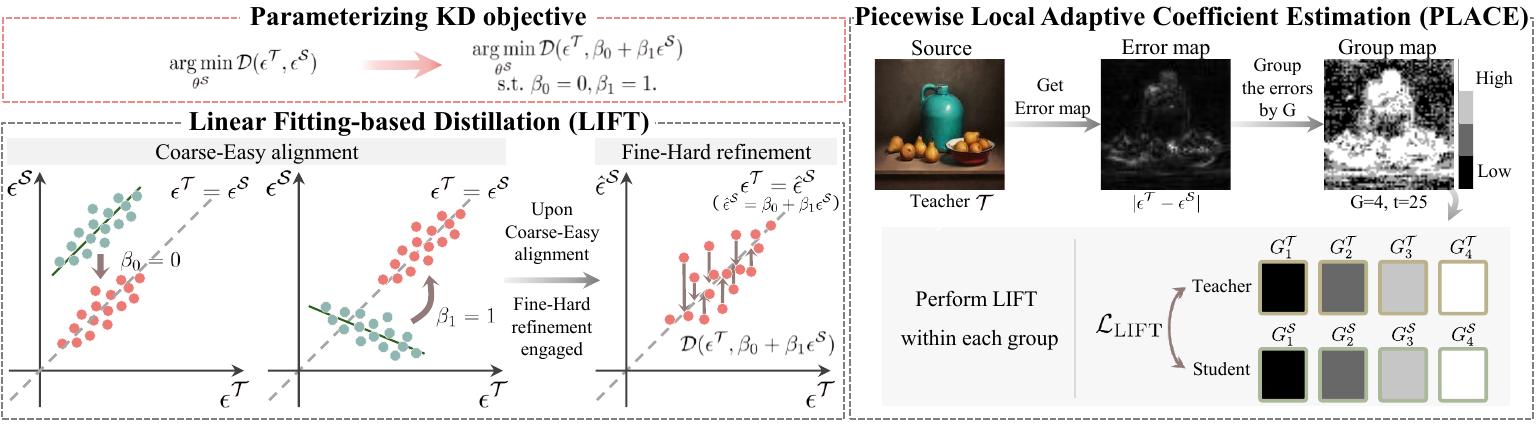}
  \caption{Overview of \lift\ and \place. LIFT parameterizes KD via linear regression, regularizing $(\beta_0\rightarrow0,\beta_1\rightarrow1)$ to align low-order moments ``Coarse–Easy'' and using the residual to learn ``Fine–Hard'' with an adaptive weight $w$. \place\ ranks error magnitudes $\mathcal{E}$, partitions outputs into equal-sized groups, estimates $(\beta_{0,i},\beta_{1,i})$ and applies \lift\ in each group for difficulty adaptive estimation.} 
  \label{fig:overview}
\end{figure*}

Beyond its statistical components, we observe that the distillation error occurs in a spatially non-uniform manner. 
Given a noisy latent $x_t$ diffused from the latent $x_0=E(I)$, encoded by the LDM encoder $E$ from an image $I$,
\cref{fig:error_map} visualizes the latent error map $\mathcal{E} =|\epsilon^{\mathcal T}-\epsilon^{\mathcal S}|\in \mathbb R^{C \times H \times W}$ alongside its decoded error map $\mathcal{D}(\mathcal{E})$, normalized and passed through the LDM decoder
This reveals that the KD error is not random noise; rather, it shows a highly structured representation.
We find that error magnitudes strongly correlate with the image's semantic components. This error map reveals that the distillation difficulty is spatially non-uniform and densely concentrated around semantically meaningful regions. 
Furthermore, comparing \cref{fig:error_map_a,fig:error_map_b} reveals that this spatial pattern is not static but evolves over the training process.
These findings motivate a strategy that is adaptive to both the spatial structure of the error and its temporal evolution during optimization.

\section{Method}
We present a \textit{Coarse-to-Fine} KD framework for lightweight diffusion models with two components, \lift~and~\place. 
The first component, LIFT, decomposes the distillation objective into ``Coarse-Easy'' and ``Fine-Hard'' errors. This enables starting by emphasizing the ``Coarse-Easy'' errors and gradually shifts toward the ``Fine-Hard'' errors to capture fine-grained teacher’s knowledge. 
The second component is PLACE, which partitions the model outputs into distinct groups and applies the LIFT objective independently to each group, enabling more granular distillation. 
Subsequent subsections elaborate on LIFT and PLACE.

\subsection{LInear FiTting-based Knowledge Distillation} \label{sec:LIFT}
Building on the decomposition introduced in \cref{sec:analysis}, we propose a \textit{Coarse-to-Fine} KD framework that first aligns the ``Coarse-Easy" errors and subsequently refines the ``Fine-Hard" residuals. We parameterize the KD objective using the intercept $\beta_0$ and slope $\beta_1$ of a linear regression. To ensure that inference does not depend on regression coefficients, we enforce the identity constraints $\beta_0{=}0$ and $\beta_1{=}1$, which correspond to directly predicting the teacher's output without relying on regression parameters. This leads to the following constrained objective:
\begin{gather}
\argmin_{\theta^{\mathcal S}}\mathcal D(\epsilon^{\mathcal T},\beta_0+\beta_1\epsilon^\mathcal S)
,\quad \text{s.t}. ~\beta_0=0,~ \beta_1=1. \label{eq:lift_obj}
\end{gather}
Since directly enforcing this constraint (\cref{eq:lift_obj}) while optimizing $\theta^\mathcal S$ is intractable, we instead treat it as a regularization term. This yields a natural decomposition of our KD objective: the regularization term corrects the coarse, moment-level mismatch, whereas the primary objective targets the remaining fine-grained residuals. 



\noindent
\textit{Coarse-Easy alignment.} For statistical moments alignment, we regularize regression coefficients (constraint of \cref{eq:lift_obj}):
\begin{gather}\mathcal L_{\text{coarse}} = |\beta_0| +|\beta_1-1|. \label{eq:coarse}\end{gather}

\noindent
\textit{Fine-Hard refinement.} We further minimize the regression residual, which captures complex details not explained by the coarse alignment (objective of \cref{eq:lift_obj}):
\begin{gather}
\mathcal L_\text{fine} = ||\epsilon^{\mathcal{T}} - (\beta_0 + \beta_1 \epsilon^{\mathcal{S}})||_2^2, \label{eq:fine}
\end{gather}
where \(\beta_0\) and \(\beta_1\) are estimated by closed-form solution of OLS (see \cref{eq:ols}).

\paragraph{Coarse-to-Fine Strategy}
We combine these loss components into a unified objective using an adaptive weighting mechanism:
\begin{gather}
\mathcal L_\text{LIFT} = \mathcal L_\text{coarse}  + w \cdot \mathcal L_\text{fine} \label{eq:lift}
\end{gather}
where the weight $w = \left(1 - \min(1, \mathcal L_\text{coarse})\right)$ is determined by the magnitude of $\mathcal L_\text{coarse}$. Early in training, $\mathcal L_\text{coarse}$ is large, so $w=0$ and the student focuses solely on correcting coarse mismatch. As coarse alignment improves and $\mathcal L_\text{coarse}{\rightarrow}~0$, $w$ increases toward 1, gradually shifting emphasis to fine-hard refinement. 
This implements a principled Coarse-to-Fine learning schedule that stabilizes early optimization and enhances final accuracy.

\begin{table*}[t]
\caption{
  Quantitative results of CelebA and LSUN-Bedroom image space diffusion. Each row grouped by student size.
 OutKD+FeatKD often underperform fine-tuning (w/o KD), while our method achieves the best performance and convergence at a pruning ratio 90\%.
 On the challenging high-resolution LSUN-Bedroom, ours at a pruning ratio 70\% even outperforms OutKD+FeatKD at a pruning ratio 50\%.
  }
\centering
\centering{
\begin{tabular}{c|cccc|cccc}
\toprule
Data & Pruning ratio & Iters & MACs & Params & Method & FID$\downarrow$ & Precision$\uparrow$ & Recall$\uparrow$  \\ \midrule
\multirow{16}{*}{\begin{tabular}[c]{@{}c@{}}CelebA \\ (64x64)\end{tabular}} 
 &\cellcolor{teacherbg}0\% &\cellcolor{teacherbg} 500k &\cellcolor{teacherbg}23.9G & \cellcolor{teacherbg}78.7M &\cellcolor{teacherbg}Teacher & \cellcolor{teacherbg}6.48 &\cellcolor{teacherbg}0.812 &\cellcolor{teacherbg}0.587 \\ \cmidrule(){2-9} 
 & \multirow{6}{*}{50\%} & \multirow{6}{*}{200k} & \multirow{6}{*}{6.0G}   & \multirow{6}{*}{19.7M}    
 & w/o KD       & 5.30 & 0.780 & \textbf{0.615} \\
 & & & & 
 & OutKD        & 5.53 & \textbf{0.800} & 0.602  \\
 & & & & 
 & FeatKD       & 5.14 & 0.773 & 0.610  \\
 & & & & 
 & OutKD+FeatKD & 5.24 & 0.797 & 0.606  \\
 & & & & 
 & Ours ($K{=}16$) & \textbf{4.93} & 0.784 & 0.612  \\ 
 \cmidrule(l){2-9} 
 & \multirow{5}{*}{70\%} & \multirow{5}{*}{240k} & \multirow{5}{*}{4.2G}   & \multirow{5}{*}{9.2M}
 & w/o KD & 6.32 & 0.779 & \textbf{0.578}  \\
 & & & & 
 & OutKD        & 6.46 & \textbf{0.805} & 0.569  \\
 & & & & 
 & FeatKD       & 6.26 & 0.766 & 0.570  \\
 & & & & 
 & OutKD+FeatKD & 6.21 & 0.795 & 0.575   \\
 & & & & 
 & Ours ($K{=}16$) & \textbf{5.97} & 0.777 & 0.577 \\ \cmidrule(l){2-9} 
 & \multirow{5}{*}{90\%} & \multirow{5}{*}{500k} & \multirow{5}{*}{0.8G}   & \multirow{5}{*}{1.3M}                       
 & w/o KD       & 223.56 & 0.135 & 0.406  \\
 & & & & 
 & OutKD        & 55.41  & 0.164 & \textbf{0.491} \\
 & & & & 
 & FeatKD       & 240.82 & 0.116 & 0.287   \\
 & & & & 
 & OutKD+FeatKD & 211.23 & 0.130 & 0.403   \\
 & & & & 
 & Ours ($K{=}16$) & \textbf{15.73} & \textbf{0.690} & 0.366   \\ 
 \midrule 
\multirow{8}{*}{\begin{tabular}[c]{@{}c@{}}LSUN\\ Bedroom\\ (256x256)\end{tabular}} 
 & \cellcolor{teacherbg}0\% &\cellcolor{teacherbg}2400k &\cellcolor{teacherbg}248.7G &\cellcolor{teacherbg}{113.7M}  &\cellcolor{teacherbg}Teacher$^{\dag}$ &\cellcolor{teacherbg}6.9 &\cellcolor{teacherbg}N/A &\cellcolor{teacherbg}N/A  \\ \cmidrule(){2-9} 
 & \multirow{4}{*}{30\%} & \multirow{4}{*}{200k} & \multirow{4}{*}{138.8G} & {\multirow{4}{*}{63.2M}} 
 & w/o KD$^{\dag}$ & 18.6 & N/A & N/A   \\
 & & & & 
 & w/o KD       & 22.98 & 0.457 & \textbf{0.605}  \\
 & & & & 
 & OutKD+FeatKD & 23.35 & 0.465 & 0.573 \\
 & & & &  
 & Ours ($K{=}16$) & \textbf{16.57} & \textbf{0.544} & 0.562 \\ \cmidrule(l){2-9}  
 & \multirow{2}{*}{50\%} & \multirow{2}{*}{500k} & \multirow{2}{*}{62.4G}  & {\multirow{2}{*}{28.5M}} 
 & OutKD+FeatKD & 69.21 & 0.201 & 0.532  \\
 & & & &
 & Ours ($K{=}16$) & \textbf{30.49} & \textbf{0.366} & \textbf{0.541}  \\ \cmidrule(l){2-9} 
 & 70\% & 700k & 52.4G & 12.6M 
 & Ours ($K{=}16$) & \textbf{37.96} & \textbf{0.326} & \textbf{0.470}  \\ \bottomrule 
\end{tabular}}
\begin{tablenotes}
      \footnotesize
      \item[*] $^{\dag}$~reported from \cite{StructuralpruningDM}. For evaluating Precision and Recall, we reproduce this experiments.
\end{tablenotes}\label{tab:MainResults}
\end{table*}
\subsection{Difficulty Adaptive Coefficient Estimation}
Building on the spatial non-uniformity analysis in \cref{sec:err}, we introduce \bplace~(\place), which retains the simplicity of LIFT while more effectively addressing heterogeneous distillation difficulty across spatial regions.
PLACE achieves this by adapting the regression coefficients to the local difficulty of each output subset, rather than enforcing a single global alignment, thereby guiding the distillation process to focus on harder regions.
We measure distillation difficulty using the error magnitude $\mathcal{E} =|\epsilon^{\mathcal T}-\epsilon^{\mathcal S}|\in \mathbb R^{C \times H \times W}$.
PLACE sorts all elements of $\mathcal{E}$ and partitions them into $C\times N$ equal-sized groups $\mathbf{G} \in \mathbb R^{C\times N\times K}=\{G_i\}_{i=1}^{C\times N}$, each containing $K$ elements. 
Equal-sized grouping keeps the method simple yet effective by avoiding complex group construction while enabling parallel coefficient estimation.
Here, $G_1$ contains the $K$ smallest errors, whereas later groups correspond to progressively harder regions.
For each group $G_i$, PLACE estimates its own regression coefficients $\beta_{0,i}, \beta_{1,i}$ using the OLS solution (\cref{eq:ols}), and computes a group-wise LIFT loss (\cref{eq:lift}).
The group size $K$ must be sufficiently large to yield stable coefficient estimates yet small enough to remain locally adaptive. 
This group-wise adaptive estimation makes the Coarse–Easy alignment term (\cref{eq:coarse}) spatially adaptive, enabling the model to better handle the varying difficulty across heterogeneous output subsets.



\noindent
\textbf{Training Objective. }
Our full KD objective integrates the diffusion loss, LIFT/PLACE loss, and feature-level KD:
\begin{align}
    \mathcal L= \lambda_{diff}\mathcal L_{diff}+ \lambda_{\text{LIFT}} \mathcal L_{\text{LIFT}}+\lambda_{\text{FeatKD}} \mathcal{L}_{\text{FeatKD}}, \label{eq:tot_loss}
\end{align}
where $\mathcal L_{diff} =||\epsilon^{\mathcal T}-\epsilon^{\mathcal S}||^2_2$. 
Under PLACE, $\mathcal L_{\text{LIFT}}$ is calculated for each group $G_i$, allowing fine-grained difficulty adaptation without additional parameters or inference cost.

\begin{table*}[t]
\caption{
Quantitative results on text-to-image diffusion models, evaluating both Stable Diffusion v2.1 (SD 2.1) and Stable Diffusion 3-Medium (SD 3). SD 3 employs an MMDiT architecture trained via flow matching, where D denotes the model depth, and its student is initialized via ShortGPT~\cite{men2025shortgpt}. MACs and Params are measured excluding the VAE and text encoders. Ours replaces only OutKD with \lift~and~\place. $^{\dag}$ indicates that FID, IS, and CLIP are reported from \cite{kim2024bk}.
}
\centering
\begin{tabular}{c|c|cccc|cccc}
\toprule
Data & Gen. Type & Architecture & Iters & MACs & Params & Method & FID$\downarrow$ & IS$\uparrow$ & CLIP$\uparrow$ \\
\midrule
\multirow{12}{*}{\begin{tabular}[c]{@{}c@{}}LAION\\ Aesthetics\\ V2 6.5+\\ (512$\times$512)\end{tabular}}

& \multirow{8}{*}{Diffusion}
& \cellcolor{teacherbg}SD 2.1
& \cellcolor{teacherbg}1620k
& \cellcolor{teacherbg}339.2G
& \cellcolor{teacherbg}865.9M
& \cellcolor{teacherbg}Teacher$^{\dag}$
& \cellcolor{teacherbg}13.93
& \cellcolor{teacherbg}35.93
& \cellcolor{teacherbg}0.3075 \\
\cmidrule(){3-10}

& & \multirow{2}{*}{BK-Base-v2}  & \multirow{2}{*}{50k} & \multirow{2}{*}{224.1G} & \multirow{2}{*}{583.5M} & BK-SDM$^{\dag}$ & \textbf{15.85} & 31.70 & 0.2868 \\
& &                              &                      &                         &                         & Ours ($K{=}16$) & 16.72 & \textbf{31.82} & \textbf{0.2900} \\
\cmidrule(l){3-10}

& & \multirow{2}{*}{BK-Small-v2} & \multirow{2}{*}{50k} & \multirow{2}{*}{218.0G} & \multirow{2}{*}{485.8M} & BK-SDM$^{\dag}$ & 16.61 & 31.73 & 0.2901 \\
& &                              &                      &                         &                         & Ours ($K{=}16$) & \textbf{14.65} & \textbf{31.99} & \textbf{0.2918} \\
\cmidrule(l){3-10}

& & \multirow{2}{*}{BK-Tiny-v2}  & \multirow{2}{*}{50k} & \multirow{2}{*}{205.3G} & \multirow{2}{*}{326.8M} & BK-SDM$^{\dag}$ & 15.68 & 31.64 & 0.2897 \\
& &                              &                      &                         &                         & Ours ($K{=}16$) & \textbf{14.60} & \textbf{31.75} & \textbf{0.2916} \\
\cmidrule(){2-10}

& \multirow{6}{*}{Flow}
& \cellcolor{teacherbg}SD 3 (D24)
& \cellcolor{teacherbg}N/A
& \cellcolor{teacherbg}799.24G
& \cellcolor{teacherbg}2.03B
& \cellcolor{teacherbg}Teacher
& \cellcolor{teacherbg}20.26
& \cellcolor{teacherbg}40.12
& \cellcolor{teacherbg}0.3089 \\
\cmidrule(){3-10}

& & \multirow{2}{*}{SD 3 (D21)} & \multirow{2}{*}{50k} & \multirow{2}{*}{699.08G} & \multirow{2}{*}{1.77B} & OutKD${+}$FeatKD & 21.17 & 36.82 & \textbf{0.3030} \\
& &                             &                      &                          &                        & Ours ($K{=}16$) & \textbf{20.59} & \textbf{37.03} & 0.3022 \\
\cmidrule(l){3-10}

& & \multirow{2}{*}{SD 3 (D18)} & \multirow{2}{*}{50k} & \multirow{2}{*}{598.83G} & \multirow{2}{*}{1.52B} & OutKD${+}$FeatKD & 22.72 & 24.28 & 0.2662 \\
& &                             &                      &                          &                        & Ours ($K{=}16$) & \textbf{21.21} & \textbf{25.14} & \textbf{0.2696} \\
\bottomrule
\end{tabular}
\label{tab:t2i_evalutations}
\end{table*}
\begin{figure*}[t]
  \centering
  \vspace{-1.0\intextsep}
    \includegraphics[width=1.0\textwidth]{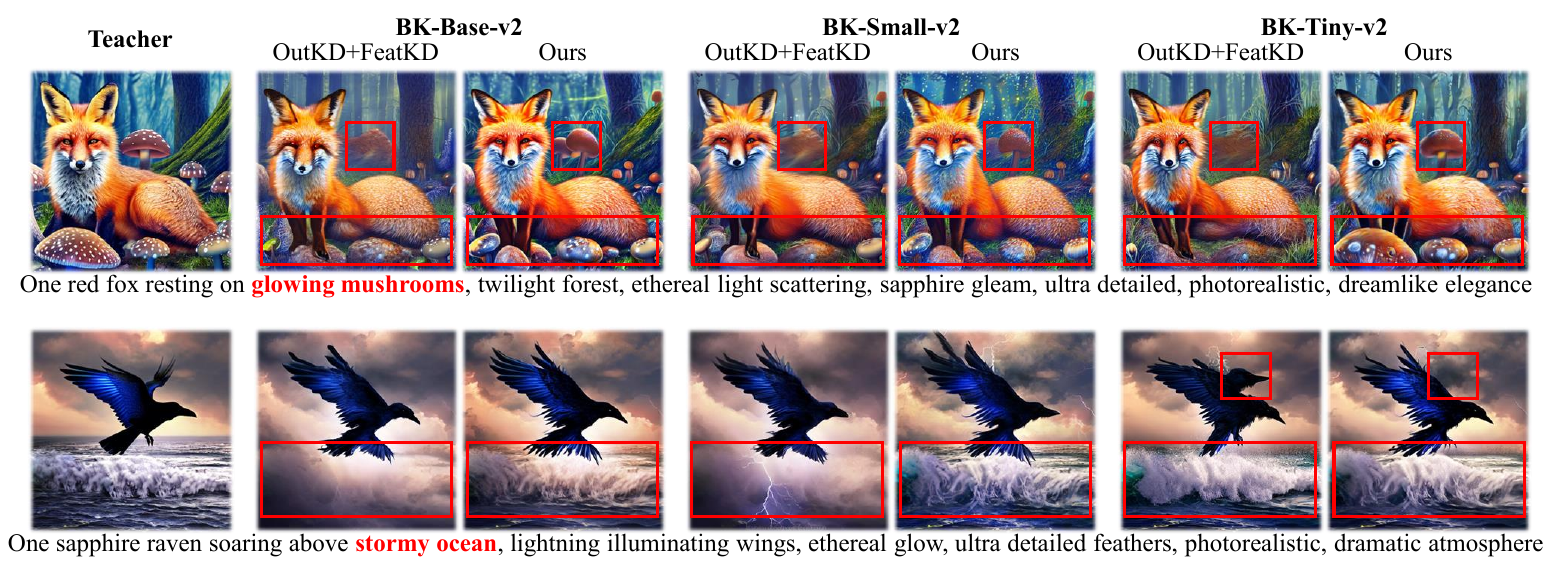}
  \caption{Qualitative results of pruned SD 2.1. Our method achieves improved semantic adherence to red-highlighted background cues.
  }
  \label{fig:t2i_qual}
\end{figure*}
\section{Experiments}
\noindent
\textbf{Student Initialization. }
Across all experiments, student models are initialized using their respective baseline pruning methods. Specifically, we employ Diff-Pruning~\citep{StructuralpruningDM} for image space diffusion; BK-SDM~\citep{kim2024bk} and SD3-Medium~\cite{esser2024scaling} (pruned via ShortGPT~\cite{men2025shortgpt}) for text-to-image latent diffusion; and TinyFusion~\citep{fang2025tinyfusion} for class-conditioned DiT. We maintain the original training setups for all baselines, replacing only OutKD with \lift~and \place.



\noindent
\textbf
{Experimental Datasets. }
\label{sec:exp_details}
We evaluate on five datasets to cover image space diffusion, text-to-image latent diffusion, and DiT. For image space diffusion, we use CelebA-HQ ($64{\times64}$) and LSUN-Bedroom ($256{\times}256$) as the low and high-resolution datasets~\citep{celeba,yu2015lsun}, respectively.
For text-to-image latent diffusion, we train on 212K samples from LAION-Aesthetics V2 6.5+~\citep{laion-aestheticsv2} and perform zero-shot evaluation on 30K MS-COCO~\citep{mscoco} validation split (following BK-SDM). For DiT experiments, we use ImageNet \citep{deng2009imagenet}.

\noindent
\textbf
{Evaluation Metrics. } 
We evaluate both computational efficiency and generative performance. For computational efficiency, we report Multiply-Add Accumulation (MACs) and the number of parameters (Params). 
For generative performance, we use Frechet Inception Distance (FID)~\citep{fid} and Inception Score (IS). To assess fidelity and diversity, we report Precision {\&} Recall~\citep{pr} .
For text-conditioned models, we use CLIP score (CLIP)~\citep{clipscore} for text-image alignment.

\subsection{Image Space Diffusion Compression} 

\begin{table*}[t]
\caption{Quantitative results on ImageNet class-conditioned DiT. Where D in architecture is depth.
  TinyFusion \citep{fang2025tinyfusion} applied block-pruning and finetuned with OutKD and masked-representation KD loss. Our model replaces OutKD loss with \textbf{LIFT} and \textbf{PLACE}, keeping the masked-representation loss. $^{\dag}$ indicates that FID, IS, Precision and Recall are reported in \cite{fang2025tinyfusion}.}
  \centering{
        \begin{tabular}{c|cccc|ccccc}
        \toprule
        Data & Architecture & Iters & MACs & Params & Method & FID$\downarrow$ & IS$\uparrow$ & Precision$\uparrow$ & Recall$\uparrow$ \\ \midrule 
        \multirow{9}{*}{\begin{tabular}[c]{@{}c@{}}ImageNet\\ (256$\times$256)\end{tabular}} 
        & \cellcolor{teacherbg} DiT-XL/2
        & \cellcolor{teacherbg}7000K & \cellcolor{teacherbg}118.7G & \cellcolor{teacherbg}675.1M & \cellcolor{teacherbg}Teacher & \cellcolor{teacherbg}2.27 & \cellcolor{teacherbg}278.2 & \cellcolor{teacherbg}0.83 &\cellcolor{teacherbg}0.57             \\ \cmidrule(){2-10} 
        & \multirow{2}{*}{DiT-D14} & \multirow{2}{*}{500k} & \multirow{2}{*}{59.4G} & \multirow{2}{*}{340.5M} 
                  & TinyFusion$^{\dag}$      & 2.86            & 234.5        & 0.82                & 0.55             \\
        & & & & & Ours ($K{=}16$) & \textbf{2.79}            & \textbf{235.6} &  0.82               & 0.55             \\ 
        \cmidrule(l){2-10} 
        & \multirow{3}{*}{DiT-D7}  & \multirow{3}{*}{500k} & \multirow{3}{*}{29.7G} & \multirow{3}{*}{173.1M} 

                & TinyFusion$^{\dag}$         & 5.87           & 166.9         & 0.78                & 0.53             \\
        & & & & & TinyFusion                  & 5.99           & 164.0        & 0.78              & 0.53            \\
        & & & & & Ours ($K{=}16$) & \textbf{5.91}           & \textbf{164.2}        & 0.78              & 0.53          \\
        \cmidrule(l){2-10}

        & \multirow{2}{*}{DiT-D4}  & \multirow{2}{*}{500k} & \multirow{2}{*}{17.0G} & \multirow{2}{*}{101.4M} 
        & TinyFusion                   & 22.11           & 78.7        & 0.61               & 0.54             \\
        & & & & & Ours ($K{=}16$)& \textbf{21.17}           & \textbf{79.5}       & 0.62              & 0.54             \\
        \bottomrule
        \end{tabular} }
    \label{tab:DiTResults}
\end{table*}
\cref{tab:MainResults} presents results for KD on image space diffusion models. At pruning ratios of 50\% and 70\%, OutKD achieves FIDs of 5.53 and 6.46 on CelebA, respectively, failing to surpass the fine-tuning (e.g., w/o KD), which attains 5.30 and 6.32. 
While FeatKD yields marginal gains, the naive combination of OutKD and FeatKD shows inconsistent performance across pruning ratios, indicating high optimization sensitivity. In contrast, our method outperforms all baselines, achieving FIDs of 4.93 and 5.97 by adaptively down-weighting hard refinement to avoid interfering with FeatKD.
Even at the extreme pruning ratio of 90\%, where conventional KD methods fail to deliver reasonable performance, our method achieves a competitive FID of 15.73. 
On LSUN-Bedroom (256${\times}$256) at a pruning ratio of 30\%, the OutKD+FeatKD achieves an FID of 23.35, which is worse than fine-tuning  at 22.98. By contrast, our method achieves an FID of 16.57, outperforming both OutKD+FeatKD and fine-tuning. At a pruning ratio of 70\%, our method attains an FID of 37.96, surpassing OutKD+FeatKD at a pruning ratio of 50\%, where it achieves an FID of 69.21, while using fewer parameters.

\subsection{Latent Space Diffusion Compression} 
Latent space diffusion models are the standard for modern high-resolution synthesis. To assess generalizability, we evaluate our framework on leading latent diffusion systems across diverse architectures and conditioning types.

\begin{figure}[t]
  \centering
  \begin{subfigure}[t]{0.49\linewidth}
    \centering
    \includegraphics[width=\linewidth]{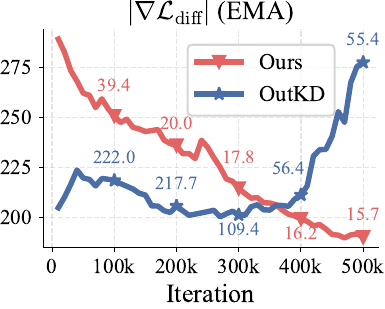}
    \captionsetup{width=1\linewidth}
    \caption{Gradient norm of diffusion loss.}
    \label{fig:gradnorm}
  \end{subfigure}
  \hfill
  \begin{subfigure}[t]{0.49\linewidth}
    \centering
    \includegraphics[width=\linewidth]{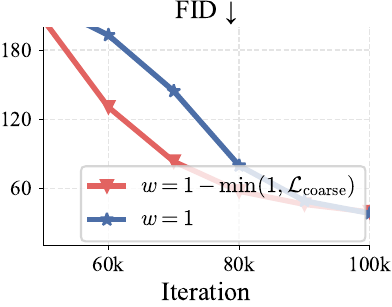}
    \captionsetup{width=1\linewidth}
    \caption{Effects of the adaptive weight.}
    \label{fig:abl_curri}
  \end{subfigure}
  \captionsetup{width=1\linewidth}
  \caption{(a) Numerical labels indicate FID at each iteration. Decreasing of gradient norm demonstrates stable convergence of our method. 
  (b) We compare our adaptive weight $w{=}1{-}\min(1,\mathcal L_\text{coarse})$ with a fixed weight $w{=}1$ that jointly optimizes ``Coarse-Easy'' and ``Fine-Hard'' errors. Prioritizing coarse information in early iterations provides more meaningful guidance and leads to faster convergence.
  }

  \label{fig:abl_all}
\end{figure}

\paragraph{Text-to-Image Generation with U-Net. }
We extend our evaluation from image diffusion to text-to-image latent diffusion. Following BK-SDM, we evaluate three student models of different sizes on SD 2.1.
In \cref{tab:t2i_evalutations}, our method consistently improves CLIP score and IS over baselines. Although FID on BK-Base-v2 slightly degrades, our method still achieves better results on the other metrics.
Qualitatively, \cref{fig:t2i_qual} shows that background-oriented prompt tokens (highlighted in red), such as \emph{glowing mushrooms} and \emph{stormy ocean}, are better captured than the baseline, yielding more faithful alignment of global scene cues.

\paragraph{Text-to-Image Generation with Flow-based Models. }
To assess our framework's generalization with flow-matching and MMDiT architectures, we extend our evaluation to SD3-Medium.
As shown in \cref{tab:t2i_evalutations}, while the D21 (depth of 21 blocks) model exhibits a marginal decrease in CLIP score, our method yields consistent improvements across all other metrics. Notably, these performance gains are particularly pronounced under the larger capacity gap, such as D18 student. These results highlight the robust generalizability of our method to flow-based models.

\paragraph{Class Conditioned Generation with DiT. }
We further evaluate our method on TinyFusion~\citep{fang2025tinyfusion} to verify whether its effectiveness extends to DiT~\citep{dit}.
This baseline uses a combination of OutKD and Masked-representation KD \citep{fang2025tinyfusion} that masks outliers of intermediate features. Because our method focuses on output-level KD, we only replace OutKD with LIFT and PLACE.
As shown in \cref{tab:DiTResults}, across multiple model depths, our method consistently improves the baseline. For the DiT-D14 model that has 14 transformer blocks, our method achieves an FID of 2.79, slightly above the TinyFusion baseline's FID of 2.86. 

\begin{table}[t]
    \caption{
    FID comparison under different $w$ schedulers. where $i$ and $I$ denote the current and total training iterations. All of student models distilled from the strongest 78.7M-teacher.
    }
    \small
    \centering
    \resizebox{1\linewidth}{!}{
    \renewcommand{\arraystretch}{1.0}
    \begin{tabular}{@{}c|ccc@{}}
        \toprule
        $w$ scheduler & 19.7M  & 9.2M & 1.3M   \\
        \midrule
        Linear ($w = i/I$) & 5.06  & 6.02 & 18.09 \\
        Cosine ($w = \frac{1}{2}(1 - \cos(\frac{i\pi}{I}))$) & 4.96  & 6.00 & 17.45 \\
        Adaptive ($w = 1 - \min(1, \mathcal{L}_{\text{coarse}})$) & \textbf{4.93}  & \textbf{5.97} & \textbf{15.73} \\
        \bottomrule
    \end{tabular}
    }
    \label{tab:abl_sch}
\end{table}




\subsection{Convergence Analysis}
We investigate the convergence behavior of compressed diffusion models. 
As illustrated in \cref{fig:gradnorm}, the gradient norm of the diffusion loss $|\nabla \mathcal L_{diff}|$ consistently decreases as training progresses. 
Prior work has reported that explicitly penalizing large gradient norms can lead to more stable optimization and better generalization \citep{zhao2022penalizing}.
Although our setup does not employ any such penalty, the observed monotonic reduction in gradient magnitude suggests that 
our method stabilizes training under aggressive compression.
This provides empirical evidence that our students, even when heavily pruned, maintain stable optimization dynamics.

\subsection{Ablation Study} \label{sec:abl}
\paragraph{Effects of Coarse-to-Fine Strategy. } 
We validate our ``Coarse-to-Fine'' strategy by comparing our adaptive weight $w = 1 - \min(1, \mathcal{L}_{\text{coarse}})$ against simultaneous optimization ($w=1$) and predefined heuristic schedulers. As shown in \cref{fig:abl_curri}, simultaneous optimization hinders early training, whereas our approach prioritizes coarse alignment to enable significantly faster convergence. Furthermore, \cref{tab:abl_sch} demonstrates that our adaptive scheduler consistently achieves the lowest FID across all student capacities compared to predefined schedulers. 
By dynamically shifting emphasis to $\mathcal L_\text{fine}$ determined by $\mathcal L_\text{coarse}$ rather than fixed training steps, our adaptive mechanism ensures a principled and optimal learning transition.

\paragraph{Effects of Teacher Capacity. } 
We investigate how the teacher-student capacity gap affects KD performance on the diffusion model. 
As shown in \cref{tab:teacher_capacity}, conventional KD degrades as the teacher's capacity increases, failing to converge when distilling from the strongest 78.7M-teacher. This indicates that exact matching becomes challenging for lightweight students. In contrast, LIFT and PLACE effectively mitigate this large capacity gap, avoiding training collapse and achieving the best, most stable performance with FID of 17.03 ${\pm}$ 1.77 with the strongest teacher. This demonstrates that our framework enables small students to reliably learn from significantly stronger teachers.

\begin{figure}
  \centering
  \includegraphics[width=1\linewidth]{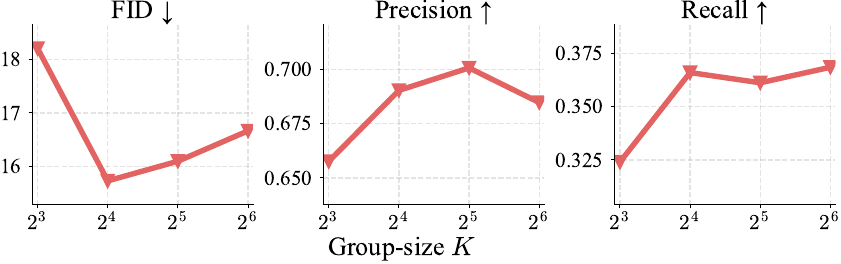}
  \captionsetup{width=1\linewidth, 
  }
  \caption{Effects of group-size $K$. Students are 90\% pruned on CelebA and distilled from the 78.7M-parameter teacher.
  }
  \label{fig:abl_groupsize}
\end{figure}

\paragraph{Effects of Group-size. }
\cref{fig:abl_groupsize} presents the FID, Precision, and Recall obtained under various group sizes $K$ in PLACE.  
When $K$ is too small ($K{=}2^3$), all metrics degrade because the group lacks sufficient samples to estimate stable regression coefficients. Conversely, when $K$ is too large ($K{=}2^6$), performance again degrades, indicating that globally estimated coefficients fail to capture the spatially non-uniform error. These findings validate that PLACE benefits from an intermediate group size that balances stability and locality.

\begin{table}[t]
\caption{Effect of teacher capacity. (90\% pruned student for CelebA). Conventional KD fails to converge or yields suboptimal results. In contrast, our method achieves the best performance and most stable convergence with the strongest teacher.}
    \centering
    \resizebox{1\linewidth}{!}{ 
    {
    \renewcommand{\arraystretch}{1.0} 
    \begin{tabular}{cc|ccc}
    \toprule
    \multicolumn{2}{c|}{\textbf{Teacher}} & \multicolumn{3}{c}{\textbf{Student (1.3M) FID}$\downarrow$}\\
    \midrule
    {Params} & {FID} & OutKD & {OutKD+FeatKD} & {Ours ($K{=}16$)} \\
    \midrule
    78.7M & 6.48 & 96.64 $\pm$ 64.99 & 193.56 $\pm$ 52.10 & \textbf{17.03 $\pm$ 1.77} \\
    19.7M & 5.30 & 78.28 $\pm$ 32.87 & 42.41 $\pm$ 6.63 & 25.38 $\pm$ 3.93 \\
     9.2M & 6.32 & 61.09 $\pm$ 23.96 & 40.88 $\pm$ 3.23 & 24.16 $\pm$ 3.48 \\
    \bottomrule
    \end{tabular}}
    }
\label{tab:teacher_capacity}
\end{table}



\section{Discussion}
\paragraph{Does higher teacher performance yield a better student in generative tasks? }
%
Standalone teacher quality does not reliably predict distillation behavior. As Tab. \ref{tab:teacher_capacity} shows, conventional KD progressively degrades as teacher capacity increases, independent of standalone performance. Although the largest teacher exhibits slightly lower quality, our method effectively distills its knowledge to achieve the best overall performance. This confirms that the largest teacher still provides highly meaningful signals, and the baseline degradation is better understood as a consequence of the large capacity gap.

\paragraph{Is there any training or inference overhead? }
Our framework simply reformulates the KD objective, introducing no additional parameters or inference overhead. Although the error-magnitude sorting in PLACE adds a resolution-dependent cost during training, the practical overhead remains negligible. In practice, the throughput drops from 4.89 to 4.86 iter/s (see \cref{sec:costs} in the Supp).

\paragraph{Generalization and Practical Advantages. }
With a negligible training overhead, our method yields comparable or improved performance across a broad range of student capacities, backbones, tasks, datasets, and diffusion spaces, and its effectiveness further extends to MMDiT and flow-based models. These results highlight a simple yet practical advantage of our \textit{Coarse-to-Fine} KD framework.

\section{Conclusion}
In this work, we investigate the capacity gap in knowledge distillation for diffusion model compression. Specifically, we observe that the complexity of knowledge makes it hard to train student models. To address this, we decompose the KD objective into ``Coarse-Easy'' and ``Fine-Hard'' alignment, and  training starts by emphasizing the “Coarse-Easy” errors and gradually shifts toward the “Fine-Hard” errors. 
Our comprehensive experiments demonstrate the effectiveness of the ``Coarse-to-Fine'' KD framework for a large capacity gap in knowledge distillation. Finally, our method successfully converges in aggressive compressed diffusion models. That pushes the limits of the diffusion model's size.

\newpage
\paragraph{Acknowledgement}
This work was supported by the National Research Foundation of Korea (NRF) grant funded by the Korea government (MSIT) (No.2022R1C1C100849612), and the Institute of Information \& Communications Technology Planning \& Evaluation (IITP) grant funded by the Korea government (MSIT) (No.RS-2020-II201336, Artificial Intelligence Graduate School Program at UNIST), (No.2022-0-00959, No.RS-2022-II220959 (Part 2) Few-Shot Learning of Causal Inference in Vision and Language for Decision Making), (RS-2025-25442149), and (RS-2025-25442824, AI Star Fellowship Program at UNIST).
We also acknowledge the high performance computing resources of the UNIST Supercomputing Center.

{
    \small
    \bibliographystyle{ieeenat_fullname}
    \bibliography{main}
}

\clearpage
\setcounter{page}{1}
\maketitlesupplementary

\begin{strip}
\centering
\centering
\captionof{table}{All of the above models use a fixed size of the student model (1.3M). For OutKD and OutKD+FeatKD, larger teachers result in higher FID means and larger FID variances.
When distilled from the strongest teacher, LIFT and PLACE achieve the lowest FID mean and standard deviation. Values in \textbf{bold} correspond to those reported in \cref{tab:MainResults}.}
\label{tab:sup_main}

\begin{tabular}{c|cc|ccccc|cc}
\toprule
Method      & Params & Teacher FID & Try 1  & Try 2  & Try 3  & Try 4  & Try 5  & \cellcolor{teacherbg}FID Mean & \cellcolor{teacherbg}FID Std \\ 
\midrule
\multirow{4}{*}{OutKD}
            & 78.7M  & 6.48        & 210.09 & 56.67  & 90.72           & \textbf{55.41} & 70.33 & \cellcolor{teacherbg}96.64    & \cellcolor{teacherbg}64.99  \\
            & 19.7M  & 5.30        & 48.84  & 88.92  & 128.96          & 50.85         & 73.84 & \cellcolor{teacherbg}78.28    & \cellcolor{teacherbg}32.87  \\
            & 16.6M  & 5.83        & 35.67  & 57.66  & 100.70          & 59.01         & 81.37 & \cellcolor{teacherbg}66.88    & \cellcolor{teacherbg}24.87  \\
            & 9.2M   & 6.32        & 41.22  & 50.48  & 82.36           & 91.14         & 40.25 & \cellcolor{teacherbg}61.09    & \cellcolor{teacherbg}23.96  \\

\midrule
\multirow{4}{*}{OutKD+FeatKD}
            & 78.7M  & 6.48        & 218.48 & 204.46 & \textbf{211.23} & 102.11 & 231.54 & \cellcolor{teacherbg}193.56   & \cellcolor{teacherbg}52.10    \\
            & 19.7M  & 5.30        & 53.68  & 40.72  & 38.41           & 42.29  & 36.94  & \cellcolor{teacherbg}42.41    & \cellcolor{teacherbg}6.63     \\
            & 16.6M  & 5.83        & 43.34  & 34.20  & 45.88           & 39.27  & 46.73  & \cellcolor{teacherbg}41.89    & \cellcolor{teacherbg}5.18     \\
            & 9.2M   & 6.32        & 44.76  & 37.82  & 43.94           & 38.44  & 39.45  & \cellcolor{teacherbg}40.88    & \cellcolor{teacherbg}3.23     \\
\midrule
\multirow{4}{*}{Ours}
        & 78.7M  & 6.48        & 19.65 & 15.80 & 18.09  & 15.85 & \textbf{15.73} & \cellcolor{teacherbg}17.03    & \cellcolor{teacherbg}1.77  \\
        & 19.7M  & 5.30        & 21.69  & 25.95 & 24.84 & 31.72 & 22.70 & \cellcolor{teacherbg}25.38    & \cellcolor{teacherbg}3.93  \\
        & 16.6M  & 5.83        & 23.50  & 20.87 & 21.09 & 29.01 & 23.24 & \cellcolor{teacherbg}23.54    & \cellcolor{teacherbg}3.28  \\
        & 9.2M   & 6.32        & 25.20  & 29.40 & 20.05 & 23.56 & 22.58 & \cellcolor{teacherbg}24.16    & \cellcolor{teacherbg}3.48  \\   
\bottomrule
\end{tabular}
\end{strip}

\section{Experimental Results of Figure 1}

\cref{fig:teaser} illustrates the challenge of the teacher--student capacity gap in KD for lightweight diffusion models. We fix a 90\%-pruned 1.3M-student and distill it from four teachers of varying capacities (78.7M, 19.7M, 16.6M, and 9.2M), evaluating two KD objectives under fixed hyperparameters. Each setting is run five times, and \cref{tab:sup_main} reports the mean and standard deviation of FID. Conventional KD becomes increasingly unstable as teacher capacity grows: both $\mathcal{L}_{\text{OutKD}}$ and $\mathcal{L}_{\text{OutKD}} + \mathcal{L}_{\text{FeatKD}}$ show worse mean FID and higher variance with larger teachers. The 78.7M-teacher gives the worst results, with most FeatKD runs collapsing.

These results suggest that the degradation of conventional KD is driven more by the teacher--student capacity gap than by the teacher's own generative quality. The difference in teacher FID is relatively small (6.48 vs.~5.30), yet the resulting student performance can collapse dramatically to FIDs of 90--200+, which is difficult to explain solely by teacher quality. Moreover, teacher FID does not reliably predict KD success: OutKD degrades more severely with the better-performing 19.7M teacher than with the 9.2M teacher, and collapses entirely with the 78.7M teacher. Similar trends are also observed in \cref{tab:DiT_abls}.

In contrast, when OutKD is replaced with LIFT and PLACE under the same setting, the strongest teacher achieves the best mean FID and the smallest standard deviation across five runs. Notably, if the 78.7M teacher provided intrinsically poor supervision, all methods should degrade similarly. Instead, our method remains stable and achieves the best mean/std with this teacher (17.03/1.77), indicating that the teacher signal itself remains useful, while naive KD becomes unstable under a large capacity gap. Overall, these results show that LIFT and PLACE mitigate the optimization difficulty caused by large capacity gaps and enable lightweight students to benefit from stronger teachers.

\begin{table*}[t]
\caption{Effects of teacher capacity in DiT. We compare the TinyFusion baseline with our method when distilling a DiT-D7 student from two teachers of different capacities. As in pixel-space diffusion, larger teachers degrade FID for TinyFusion, whereas our LIFT and PLACE provide consistent improvements, with a slightly larger gain when distilling from the stronger teacher. \textbf{Bold} indicates the best performance for each teacher used in distillation, and $^{\dag}$ denotes metrics reported in \cite{fang2025tinyfusion}.}
  \centering{
        \begin{tabular}{c|cccc|ccccc}
        \toprule
        Data & Architecture & Iters & MACs & Params & Method & FID$\downarrow$ & IS$\uparrow$ & Precision$\uparrow$ & Recall$\uparrow$ \\ \midrule 
        \multirow{8}{*}{\begin{tabular}[c]{@{}c@{}}ImageNet\\ (256$\times$256)\end{tabular}} 
    
        & \cellcolor{teacherbg} DiT-D14
        & \cellcolor{teacherbg}500K & \cellcolor{teacherbg}59.4G & \cellcolor{teacherbg}340.5M & \cellcolor{teacherbg}Teacher & \cellcolor{teacherbg}2.86 & \cellcolor{teacherbg}234.5 & \cellcolor{teacherbg}0.82 &\cellcolor{teacherbg}0.55             \\

        & \multirow{3}{*}{DiT-D7}  & \multirow{3}{*}{500k} & \multirow{3}{*}{29.7G} & \multirow{3}{*}{173.1M} 

                & TinyFusion$^{\dag}$         & 5.87           & 166.9         & 0.78                & 0.53             \\
        & & & & & TinyFusion                  & 5.99           & 164.0        & 0.78              & 0.53            \\
        & & & & & Ours ($K{=}16$) & \textbf{5.91}           & \textbf{164.2}        & 0.78              & 0.53          \\
        \cmidrule(l){2-10}  
        
        & \cellcolor{teacherbg} DiT-XL/2
        & \cellcolor{teacherbg}7000K & \cellcolor{teacherbg}118.7G & \cellcolor{teacherbg}675.1M & \cellcolor{teacherbg}Teacher & \cellcolor{teacherbg}2.27 & \cellcolor{teacherbg}278.2 & \cellcolor{teacherbg}0.83 &\cellcolor{teacherbg}0.57             \\ 
         
        & \multirow{2}{*}{DiT-D7} & \multirow{2}{*}{500k} & \multirow{2}{*}{29.7G} & \multirow{2}{*}{173.1M} 
                  & TinyFusion      & 7.87            & 145.5        & 0.75                & 0.54             \\

        & & & & & Ours ($K{=}16$) & \textbf{7.66}            & \textbf{146.9} &  0.75               & 0.53             \\ 
        


        \bottomrule
        \end{tabular} }
    \label{tab:DiT_abls}
\end{table*}

\section{Experiments Details}
Across all experiments, we fix PLACE’s group size to $K{=}16$, as determined by our ablation study (see \cref{fig:abl_groupsize}). For image space diffusion models, we use the Diff-Pruning base pruned model with varying pruning ratios, where the pruning ratio denotes the fraction of teacher channels removed. We set $\lambda_{\text{diff}}{=}1$ and $\lambda_{\text{FeatKD}}{=}1\mathrm{e}{-6}$ for all such experiments. Our main comparison is between the baseline output-level KD (OutKD) and our LIFT loss; by default, we assign equal weights, $\lambda_{\text{OutKD}}{=}1.0$ and $\lambda_{\text{LIFT}}{=}1.0$. An exception arises on CelebA with 50\% and 70\% pruning, where the student trained without KD already outperforms the teacher. For these two settings only, we reduce the guidance weight for both methods to $0.1$ (i.e., $\lambda_{\text{OutKD}}{=}0.1$ and $\lambda_{\text{LIFT}}{=}0.1$) and relax $\mathcal{L}_{\text{coarse}}$ to use an $\ell_2$ distance. For latent space diffusion models, we follow the BK-SDM and TinyFusion baselines and simply replace their output-level distillation loss $\mathcal{L}_{\text{OutKD}}$ with our $\mathcal{L}_{\text{LIFT}}$. All other hyperparameters, including their respective $\lambda$ weights, are kept identical to the original papers for a fair comparison. This also includes adopting TinyFusion’s schedule, which anneals the feature-level KD weight $\lambda_{\text{MaskedFeatKD}}$ to $0$ during training.

\begin{table}[t]
    \caption{Ablation study on the components of \cref{eq:tot_loss}. All of student models distilled from the strongest 78.7M-teacher.}
    \centering
    \resizebox{0.75\linewidth}{!}{
    \begin{tabular}{ccc|cc}
        \toprule
        \multicolumn{3}{c|}{{Components of \cref{eq:tot_loss}}} & \multicolumn{1}{c}{{Student (1.3M)}}\\
        \midrule
        {LIFT} & {PLACE} & {FeatKD} & FID$\downarrow$ \\
        \midrule
        \checkmark & \checkmark & \checkmark & \textbf{15.73} \\
        \checkmark &            & \checkmark & 18.49 \\ 
        \checkmark & \checkmark &            & 16.67 \\ 
        \checkmark &            &            & 19.07 \\  
        \bottomrule
    \end{tabular}
    }
    \label{tab:sch_ablation}
\end{table}

\section{Non-uniform Error Across Architectures}
\begin{figure}[t]
  \centering
  
  \begin{subfigure}{1.0\linewidth}
    \centering
    \includegraphics[width=\linewidth]{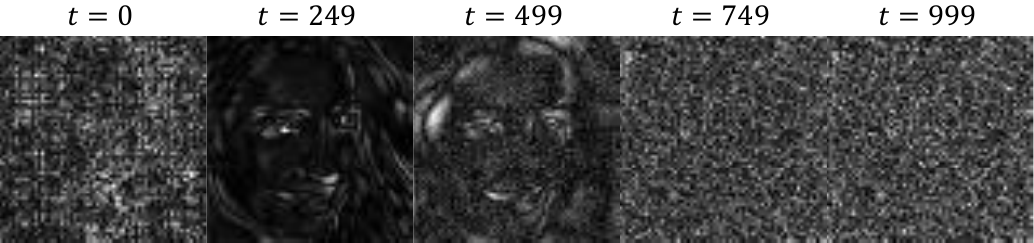}
    \captionsetup{width=1\linewidth}
    \caption{Error map of OutKD}
    \label{fig:sup_out_err}
  \end{subfigure}


  \begin{subfigure}{1.0\linewidth}
    \centering
    \includegraphics[width=\linewidth]{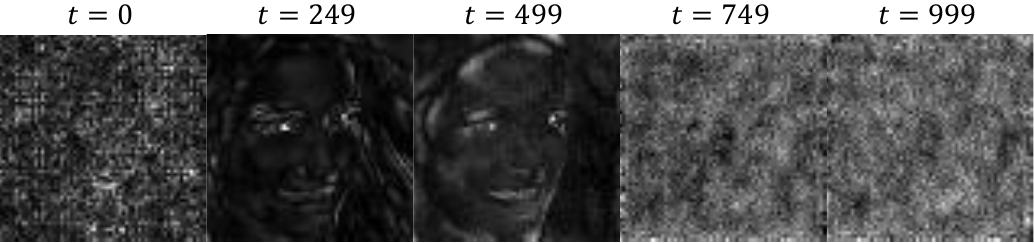}
    \captionsetup{width=1\linewidth}
    \caption{Error map of OutKD+FeatKD}
    \label{fig:sup_feat_err}
  \end{subfigure}

  \captionsetup{width=1\linewidth}
  \caption{Error map of lightweight student models after being trained by using conventional KD: (a) using only $\mathcal L_{\text{OutKD}}$, (b) using $\mathcal L_{\text{OutKD}}+\mathcal L_{\text{FeatKD}}$. We observe that the distillation error is non-uniform across diffusion time steps.}
  \label{fig:sup_ddpm_err}
\end{figure}

\begin{figure}[t]
  \centering
  \begin{subfigure}[t]{0.32\linewidth}
    \centering
    \includegraphics[width=\linewidth]{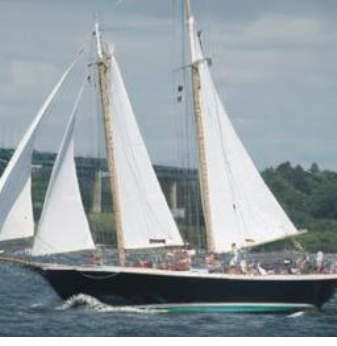}
    \captionsetup{width=1\linewidth}
    \caption{Real image}
    \label{fig:sup_dit_real}
  \end{subfigure}
  \hfill
  \begin{subfigure}[t]{0.32\linewidth}
    \centering
    \includegraphics[width=\linewidth]{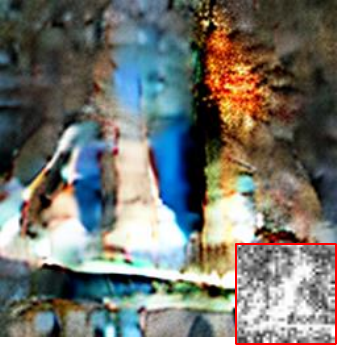}
    \captionsetup{width=1\linewidth}
    \caption{Early iteration, 5k}
    \label{fig:sup_dit_early_err}
  \end{subfigure}
  \hfill
  \begin{subfigure}[t]{0.32\linewidth}
    \centering
    \includegraphics[width=\linewidth]{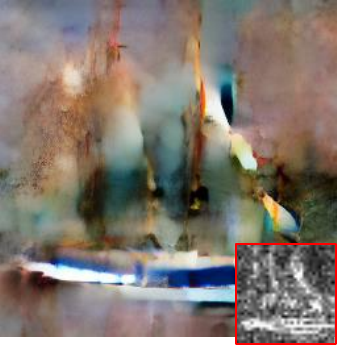} 
    \captionsetup{width=1\linewidth}
    \caption{After training, 50k}
    \label{fig:sup_dit_after_err}
  \end{subfigure}

  \captionsetup{width=1\linewidth}
  \caption{Visualization of error map of TinyFusion (i.e., DiT-D7): (a) real image, (b) early iteration, and (c) after training. Consistent with \cref{fig:error_map}, the DiT architecture also exhibits spatially non-uniform distillation error.}
  \label{fig:sup_dit_err}
\end{figure}

In \cref{sec:err} of the main paper, we showed that the distillation error between teacher and student is spatially non-uniform and exhibits a highly structured pattern that correlates with semantic content. Here, we examine whether this phenomenon persists across diffusion models with substantially different architectures.
\cref{fig:sup_ddpm_err} presents the student-teacher error map for a pixel space diffusion model.
Consistent with \cref{fig:error_map}, the distillation error is not random or uniformly distributed in other models. \cref{fig:sup_ddpm_err} shows that the distillation error concentrates on semantically meaningful regions such as facial features at the middle time step.
This confirms that spatially non-uniform distillation difficulty is not specific to latent space models, but also arises in pixel space diffusion models. 
We further provide the error map of TinyFusion. 
As shown in \cref{fig:sup_dit_err}, the DiT model is consistent with \cref{sec:err}. The error map again forms a distinct, structured pattern that aligns with semantic content.


\section{Additional Ablation Studies} \label{sec:additional_abl}
We provide detailed ablation studies to validate our LIFT and PLACE. All student models in these experiments are distilled from the strongest 78.7M-teacher model. While the CelebA experiments in \cref{tab:MainResults} (i.e., 19.7M- and 1.3M-student models) reveal that OutKD+FeatKD occasionally leads to performance degradation compared to FeatKD alone, the results in \cref{tab:sch_ablation} demonstrate that LIFT effectively stabilizes this combination. Consistent with prior literature \cite{kim2024bk, kim2025random, fang2025tinyfusion}, LIFT ensures robust performance gains. Notably, the multi-set coefficient estimation via PLACE enhances the reliability of LIFT’s optimization, resulting in a substantial performance gain in FID from 19.07 to 16.67.


\subsection{Training Overhead}
\label{sec:costs}
During training, error-magnitude sorting in PLACE introduces a resolution-dependent cost. However, as shown in \cref{tab:speed_comparison}, training throughput (iter/s) and VRAM usage on an Intel Xeon 6740P CPU and an NVIDIA RTX PRO 6000 GPU show that this overhead is negligible, since the overall cost is dominated by the model's forward and backward passes. At inference time, our method incurs no additional cost, consistent with prior works~\citep{StructuralpruningDM,kim2024bk,fang2025tinyfusion}.

\section{Related Works}
\subsection{Efficient Diffusion Model}
Although diffusion models \citep{ho2020denoising,song2020score,sd1.4,dit} demonstrate outstanding performance, their inherent iterative denoising process not only demands substantial computational resources but also makes it challenging to apply existing compression methods designed for feed-forward networks~\citep{liu2021content,chung2024diversity,yeo2024nickel,kim2025singular,xu2022mind}. To address these problems, compression methods were explored that take the unique characteristics of diffusion models. 
For fast sampling, some prior work ~\citep{song2023consistency,kim2023consistency,gu2023boot,yin2024improved,yin2024one,salimans2022progressive,song2024sdxs}~studied KD to reduce the number of denoising steps. However, this method did not consider compressing the network architecture.
In contrast, pruning has been considered for compressing diffusion models.  For instance, Diff-pruning~\citep{StructuralpruningDM} used Taylor scores to assign weight importance for pruning, accounting for varying noise levels across diffusion steps.
TinyFusion~\citep{fang2025tinyfusion} proposed block pruning and feature-level distillation by learning pruning masks and masking outliers of intermediate features.  
Moreover, KD for compressed diffusion models, such as random-conditioning~\citep{kim2025random} and DKDM~\citep{xiang2025dkdm} were proposed for data-efficient KD. In this work, we focus on KD itself for lightweight diffusion model.


\subsection{Capacity Gap in Knowledge Distillation}
Knowledge distillation, which trains a student model to mimic the teacher model's outputs and intermediate features, is a widely used approach for model compression across various tasks. By distilling the knowledge, the student model can learn the teacher’s complex knowledge with a smaller model capacity.
In various domains, the capacity gap between student and teacher models hinders the transfer of complex knowledge~\citep{mirzadeh2020improved,cho2019efficacy,huang2022knowledge,son2021densely,wang2022efficient}.
To address this problem, DIST \citep{huang2022knowledge} relaxed the matching objective by maximizing the Pearson correlation between teacher and student outputs, while TAKD \citep{mirzadeh2020improved} employed a teaching assistant with intermediate capacity to progressively bridge the gap and transfer knowledge across multiple stages.


\begin{table}[t]
    
\caption{Training overhead and peak VRAM usage evaluated on an Intel Xeon 6740P CPU and an NVIDIA RTX PRO 6000 GPU.}
    \centering
    \resizebox{1\linewidth}{!}{ 
    \renewcommand{\arraystretch}{1.0}
        \begin{tabular}{c|cc|cc}
            \toprule
            \multirow{2}{*}{\begin{tabular}[c]{@{}c@{}}Dataset\\(Resolution)\end{tabular}} &
            \multicolumn{2}{c|}{OutKD+FeatKD} & \multicolumn{2}{c}{Ours ($K{=}16$)} \\ 
            \cmidrule(){2-5}
            & Speed (iter/s) & VRAM (GB) & Speed (iter/s) & VRAM (GB) \\ \midrule
            LSUN (256$\times$256)    & 4.8935 iter/s & {5.615 GB} & 4.8634 iter/s & {5.625 GB}  \\
            LSUN (512$\times$512)    & 3.6337 iter/s & {18.287 GB} & 3.6291 iter/s & {18.326 GB} \\ 
            \bottomrule
        \end{tabular}
    }
    \label{tab:speed_comparison}
\end{table}

\section{Visualization Results}
\begin{figure*}[t]
  \centering
    \includegraphics[width=1.0\textwidth]{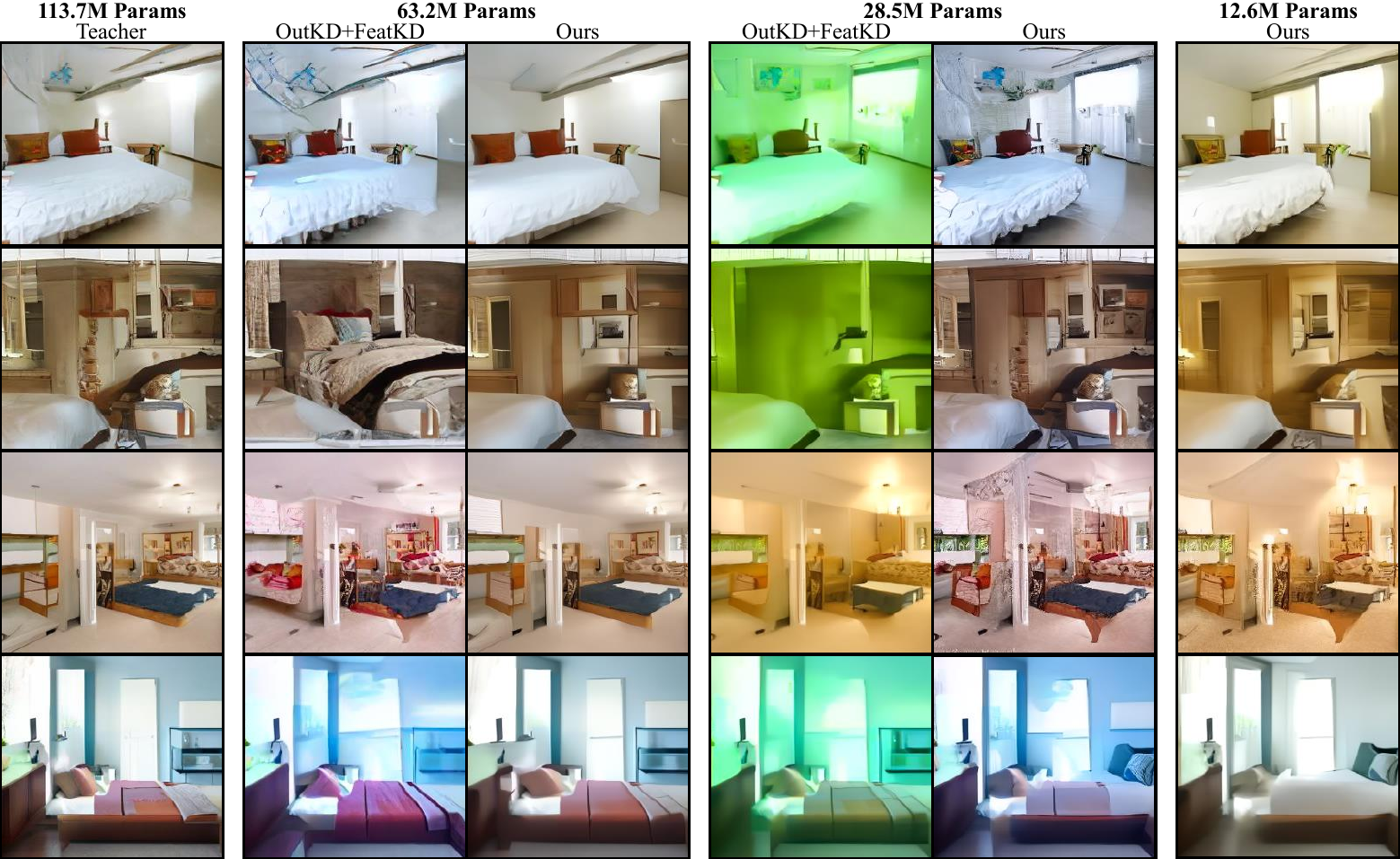}
  \caption{Visualization of pixel space diffusion models with LSUN Bedroom.
  }
  \label{fig:sup_lsun_qual}
\end{figure*}

\begin{figure*}[t]
  \centering
    \includegraphics[width=1.0\textwidth]{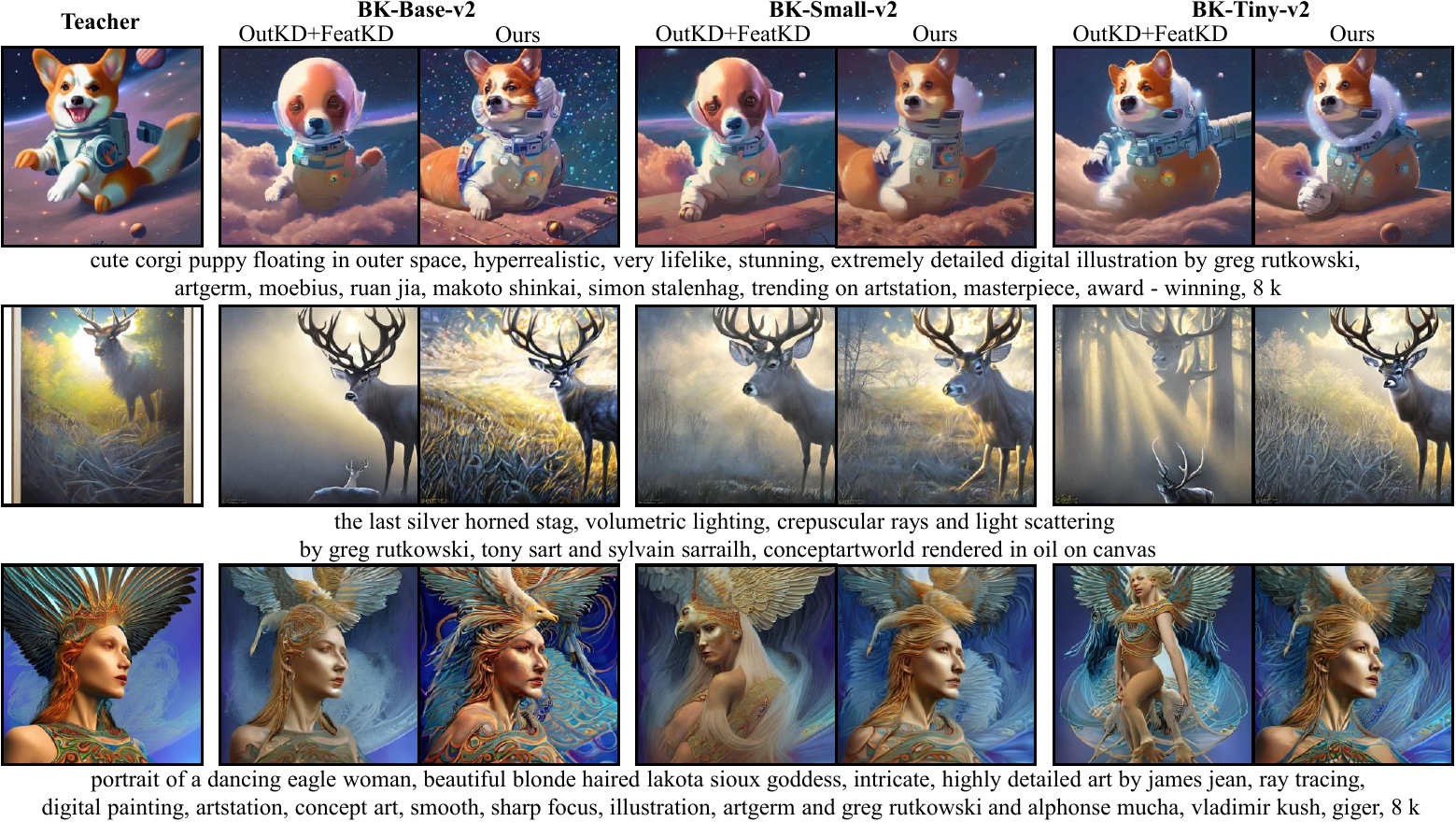}
  \caption{Visualization of pruned Stable Diffusion 2.1.
  }
  \label{fig:sup_t2i_qual}
\end{figure*}

\begin{figure*}[t]
  \centering

  \begin{subfigure}[t]{0.49\linewidth}
    \centering
    \includegraphics[width=\linewidth]{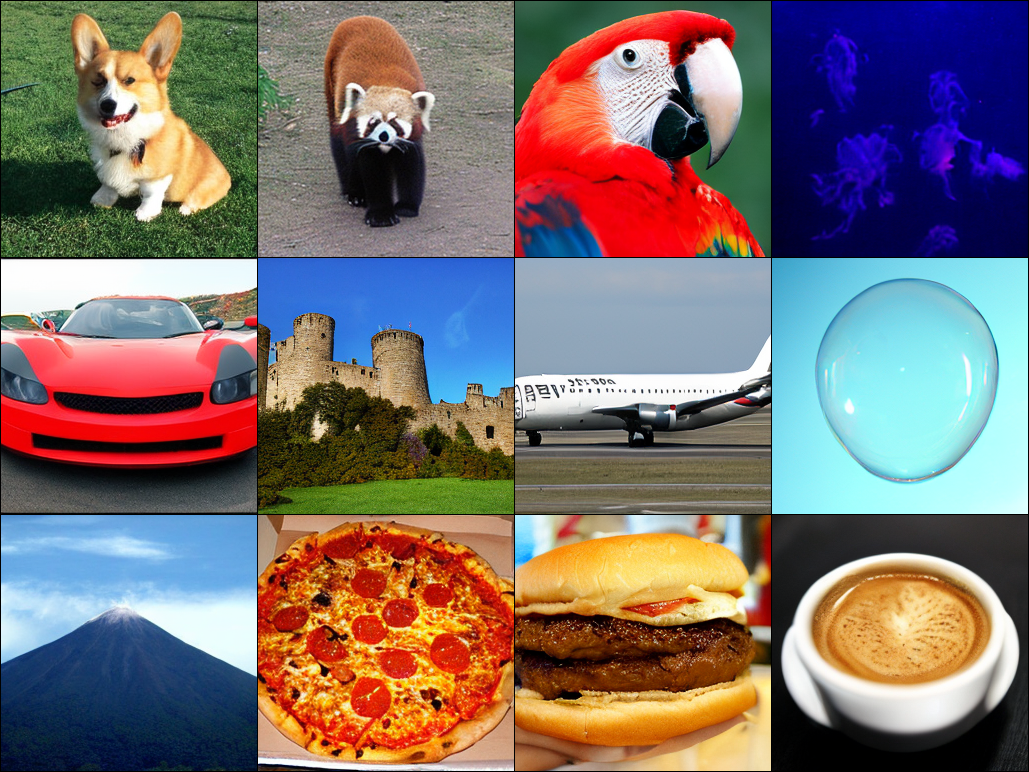}
    \caption{TinyFusion DiT-D14.}
    \label{fig:sup_dit_d14}
  \end{subfigure}
  \hfill
  \begin{subfigure}[t]{0.49\linewidth}
    \centering
    \includegraphics[width=\linewidth]{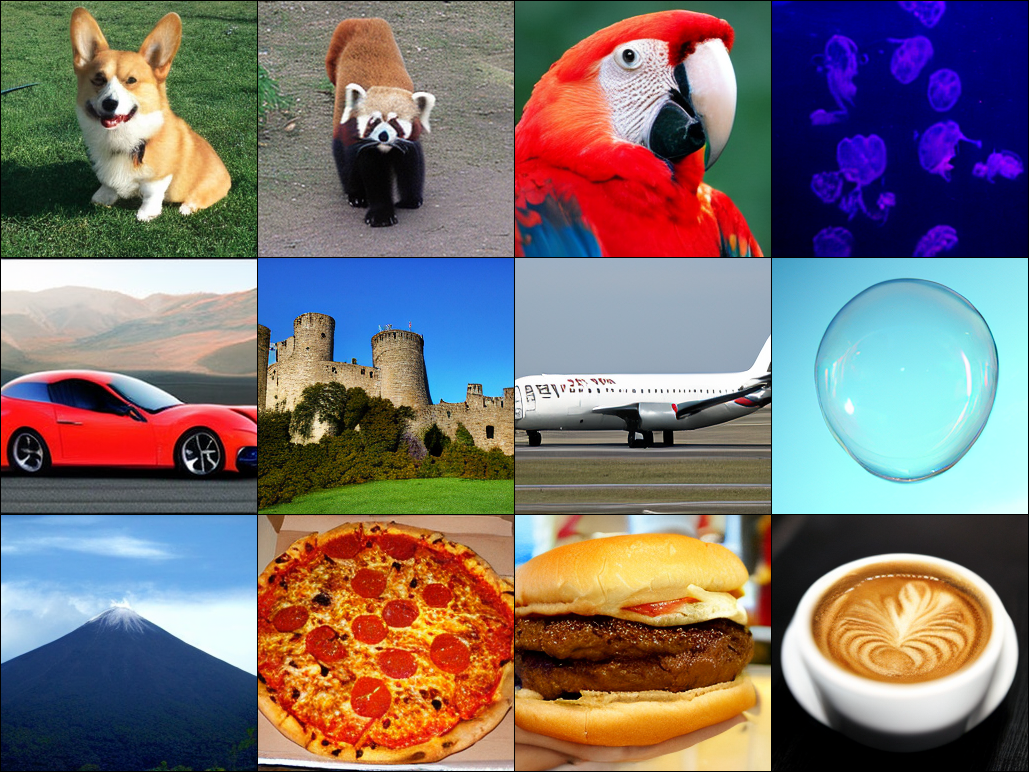}
    \caption{DiT-D14 with our method.}
    \label{fig:sup_dit_ours_d14}
  \end{subfigure}

  \vspace{0.5em}

  \begin{subfigure}[t]{0.49\linewidth}
    \centering
    \includegraphics[width=\linewidth]{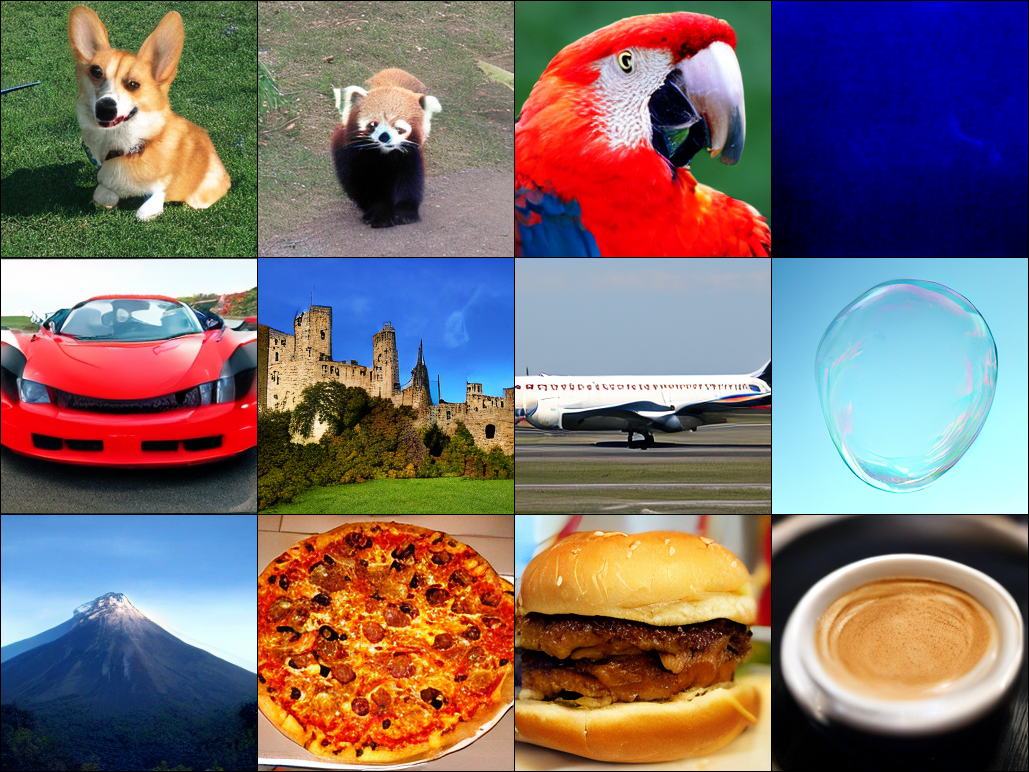}
    \caption{TinyFusion DiT-D7.}
    \label{fig:sup_dit_tiny_d7}
  \end{subfigure}
  \hfill
  \begin{subfigure}[t]{0.49\linewidth}
    \centering
    \includegraphics[width=\linewidth]{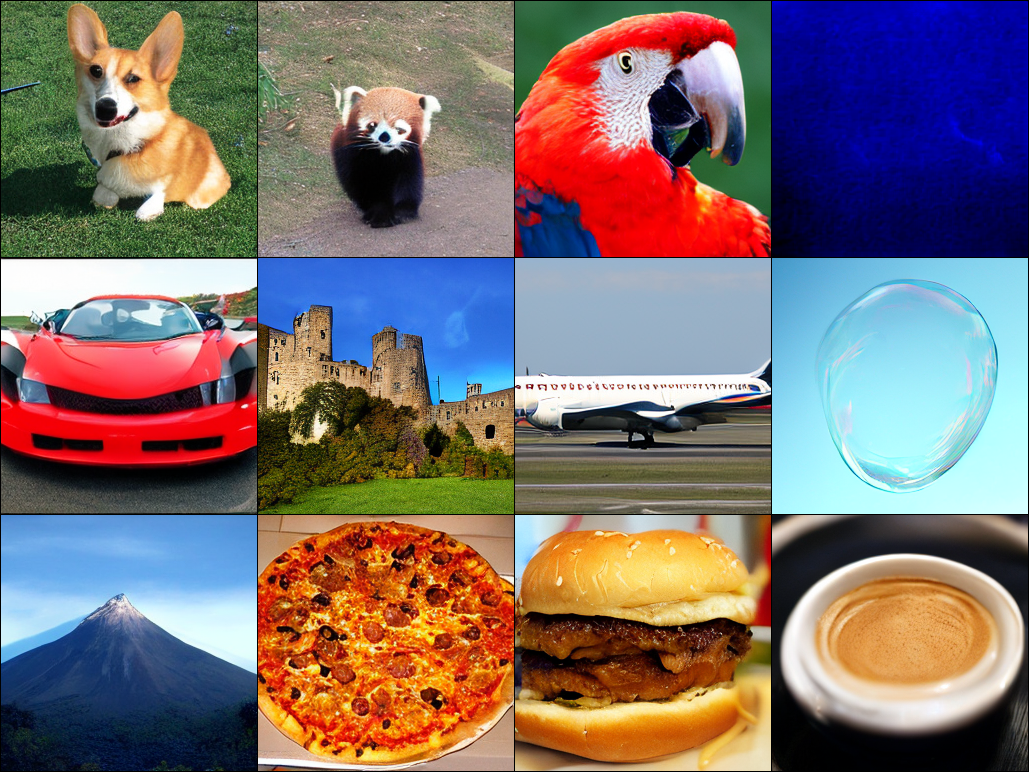}
    \caption{DiT-D7 with our method.}
    \label{fig:sup_dit_ours_d7}
  \end{subfigure}

  \caption{Visualization results for DiT-D14 and DiT-D7. The top row compares DiT-D14, and the bottom row compares DiT-D7.}
  \label{fig:sup_dit_qual}
\end{figure*}

We provide visualization results of our experiments. The following figures present representative samples for image and latent space diffusion models.
Each \cref{fig:sup_lsun_qual,fig:sup_t2i_qual,fig:sup_dit_qual} correspond to \cref{tab:MainResults,tab:t2i_evalutations,tab:DiTResults}, respectively.
The results show that across all model sizes (see \cref{fig:sup_lsun_qual}), our method produces noticeably more stable and realistic samples than OutKD+FeatKD.
\cref{fig:sup_t2i_qual} shows that our method captures not only the main subject but also details (e.g., background) compared to OutKD+FeatKD. As shown in \cref{tab:t2i_evalutations}, the overall outputs are similar to TinyFusion, while our method sometimes captures finer details more effectively.




\clearpage



\end{document}